\documentclass[journal]{IEEEtran}

\usepackage{cite}
\usepackage{amsmath,amssymb,amsfonts}
\usepackage{algorithmic}
\usepackage{graphicx}
\usepackage{textcomp}
\usepackage{xcolor}
\usepackage{booktabs}
\usepackage{multirow}
\usepackage{array}
\usepackage{threeparttable}
\usepackage{stfloats}
\usepackage{xcolor}

\newcolumntype{L}[1]{>{\raggedright\arraybackslash}p{#1}}
\newcolumntype{C}[1]{>{\centering\arraybackslash}p{#1}}
\usepackage{caption}

\usepackage{pifont}

\newcommand{\cmark}{\textcolor{green!60!black}{\ding{51}}} % green tick
\newcommand{\xmark}{\textcolor{red!70!black}{\ding{55}}}   % red cross

\begin{document}

\title{A Depth-Aware Comparative Study of Euclidean and Hyperbolic Graph Neural Networks on Bitcoin Transaction Systems}

\author{\IEEEauthorblockN{Ankit Ghimire\textsuperscript{1}, Saydul Akbar Murad\textsuperscript{1}, and Nick Rahimi\textsuperscript{1}}\\
\IEEEauthorblockA{\textsuperscript{1}\textit{School of Computing Sciences and Computer Engineering}\\
\textit{University of Southern Mississippi}
Hattiesburg, MS, 39406, USA\\
\{ankit.ghimire, saydulakbar.murad, nick.rahimi\}@usm.edu}
}

\maketitle

% =============================
% Abstract
% =============================
\begin{abstract}

Bitcoin transaction networks are large scale socio-technical systems in which activities are represented through multi-hop interaction patterns. Graph Neural Networks(GNNs) have become a widely adopted tool for analyzing such systems, supporting tasks such as entity detection and transaction classification. Large-scale datasets like Elliptic have allowed for a rise in the analysis of these systems and in tasks such as fraud detection. In these settings, the amount of transactional context available to each node is determined by the neighborhood aggregation and sampling strategies, yet the interaction between these receptive fields and embedding geometry has received limited attention. In this work, we conduct a controlled comparison of Euclidean and tangent-space hyperbolic GNNs for node classification on a large Bitcoin transaction graph. By explicitly varying the neighborhood while keeping the model architecture and dimensionality fixed, we analyze the differences in two embedding spaces. We further examine optimization behavior and observe that joint selection of learning rate and curvature plays a critical role in stabilizing high-dimensional hyperbolic embeddings. Overall, our findings provide practical insights into the role of embedding geometry and neighborhood depth when modeling large-scale transaction networks, informing the deployment of hyperbolic GNNs for computational social systems.

\end{abstract}

\begin{IEEEkeywords}
Bitcoin Transaction Networks, Graph Neural Networks, Hyperbolic Embeddings, Illicit Behavior Detection
\end{IEEEkeywords}

% =============================
% 1. Introduction
% =============================
\section{Introduction}

Bitcoin transaction networks are large-scale socio-technical systems formed by interactions among individuals, services, and organizations operating on a decentralized financial infrastructure. With Bitcoin’s substantial market capitalization, these networks attract both legitimate usage and illicit use. This has resulted in great interest in the analysis of such financial systems with the aim of detecting patterns and illicit activities such as money laundering, fraud, ransomware, and Ponzi schemes \cite{ELLIPTIC, BITCOINFRAUD1, BITCOINFRAUD2}. Understanding transactional behavior at scale is critical for maintaining the security and integrity of blockchain-based financial systems.

Bitcoin transactions are naturally represented as directed graphs which represent the direction of flow of funds through different addresses. Hence graph-based representations provide a natural framework for modeling transaction data, capturing relational structure and behavioral patterns. To analyze such graph-based representations, Graph Neural Networks (GNNs) are perfectly suited and provides an effective tool in tasks such as address entity classification, and anomaly detection on large-scale datasets such as Elliptic \cite{ELLIPTIC} and Bitcoin Orbital \cite{ORBITAL}. In these settings, GNNs have been applied to the full transaction graph since they were of size feasible for computation on the entire graph. 

However, to effectively model the transactional behavior in a Bitcoin network we need to incorporate richer multi-hop transactional context. In a computational system like Bitcoin, illicit activities are rarely characterized by isolated interactions. Instead behaviors like money laundering, ransomware payments, and Ponzi schemes typically involve coordinated sequences of transactions, and reuse of intermediary addresses. Hence, capturing such patterns requires going beyond the immediate neighbors across multiple hops, as local transaction features alone might be insufficient to distinguish different entity types and to detect illicit activities.

As the transactional context is expanded, the multi-hop transaction patterns, when traced from a source address, often exhibit structured patterns and flow of fund behaviors that resemble locally hierarchical or tree-like expansion. Dynamic analyses that trace taints have observed that the entities have a characteristic flow and the entities can be tracked or classified based on the flow of their coins\cite{tovanich2022pattern}. In laundering, ransomware and mixing activities, funds and transactions are often fragmented in multiple addresses creating fan out transactions and branching \cite{zola2025topological}. While Bitcoin transactions graphs are not hierarchical by construction, these repeated multi-step patterns and branching suggests a tree-like structure as the hop depth increases.

If such hierarchical expansion is present, then representing such patterns might be difficult for Euclidean embedding spaces. As the neighborhood depth increases, the node representations must capture information of large and branching neighborhoods inside a fixed dimensional embedding space which can create distortions. Hyperbolic geometry, on the other hand, provides a natural geometric alternative for representing such expansion patterns.

While Euclidean GNNs have demonstrated strong performance, the interaction between neighborhood depth and embedding geometry has received limited systematic investigation in Bitcoin transaction networks. Prior work generally applies GNNs to full graphs or fixes the nominal receptive field through network architecture \cite{DECOUPLE, GRAPHSAINT}, without explicitly examining how increasing sampled neighborhood depth influences representation quality under different geometric spaces. The absence of systematic comparisons between GNNs operating in different embedding spaces makes it difficult to identify effective geometric representations for Bitcoin transaction data.

Moreover, increasing the amount of aggregated transactional context introduces new challenges because as the neighborhood depth grows, the node representations must encode increasingly branching and heterogeneous patterns. Also increasing depth is known to naturally rise in oversmoothing in standard GNN architectures which limits the depth of GNN architectures in most cases. 

Hyperbolic embedding spaces have been shown to provide benefits in capturing representations of data with latent tree-like or branching structure \cite{nickel2017poincare}. Recent work has extended hyperbolic geometry to graph neural networks, demonstrating potential benefits in graph structures with hierarchical characteristics \cite{CHAMIGCN}. But whether these benefits extend to large-scale transaction graphs where branching patterns emerge with multi-hop expansion remains an open question.

In this work, we systematically evaluate whether this geometric compatibility translates into measurable performance gains in Bitcoin transaction classification. Our results show that hyperbolic GNNs consistently outperform their Euclidean counterparts, with advantages most pronounced in deeper architectures and stable optimization achievable through careful joint selection of curvature and learning rate.
The contribution of our research are as follows:
\begin{itemize}
\item We present a controlled comparison of Euclidean and tangent-space hyperbolic GNNs for node classification on subgraphs derived from a large Bitcoin transaction graph.
\item We analyze the effect of neighborhood sampling depth on model performance across different embedding spaces and also provide per-class analysis.
\item We provide quantitative evidence that hyperbolic GNNs perform better especially with deeper architectures and provide practical insights into the selection of curvature and learning rates.
\end{itemize}

The remainder of this research article is organized as follows: Section II provides an overview of related work, highlighting key studies and their implementations in different fields. Section III outlines the methodology, from the dataset used, subgraphs creation, and data processing, to the GNNs architectures that were used. Section IV presents and discusses the experimental results and the computational requirements. Section V summarizes the findings, and section VI discusses the limitations and potential directions for future work.

% =============================
% 2. Related Work
% =============================
\section{Related Work}
This section reviews prior work along three complementary dimensions relevant to our study: (i) analysis of Bitcoin transaction networks, (ii) graph neural networks for transaction and social systems, and (iii) hyperbolic representation learning for graph-structured data.
\subsection{Bitcoin Transaction and Financial Networks}
In recent years, cryptocurrencies have been widely adopted as a payment method and among them all, Bitcoin has been the most well-known and widespread. Due to the feature of anonymity, Bitcoin has become a preferred method for criminals to involve themselves in illegal activities while hiding their identity and location. This anonymity  along with the transparency of the Bitcoin transactions has generated great interest in the financial community in the analysis of Bitcoin transaction graphs with the goal of detecting illicit financial activities. Early work on Bitcoin graph analysis focused on heuristic-based tracing and graph analytics to detect suspicious transaction patterns like in  money laundering, mixing, and ransomware payments. These studies employed properties like transaction flows, address reuse, clustering based on transactions and transaction statistics to uncover coordinated behaviors\cite{weber2019anti, tovanich2022pattern}.

With the release of labeled datasets such as Elliptic and Elliptic++ \cite{ELLIPTIC, elmougy2023demystifying}, machine learning approaches became increasingly prevalent. These datasets enabled supervised classification of Bitcoin addresses and entities and motivated the application of various learning based models for tasks like fraud detection and entity classification. Recent studies have also explored the temporal and structural dependencies in these graphs for anti-money laundering applications \cite{TEMPORAL}. Most recently, large-scale reconstructions of the Bitcoin transaction graph \cite{schnoering2025bitcoin} have expanded the scope of analysis by providing millions of nodes and edges along with classified entity labels. Due to the large scale of these graphs, it allows for a long range dependency modeling and flow-of-funds analysis across entities while still allowing for localized analysis by methods like creating ego-centric subgraphs or neighborhood sampling.

Several studies report that illicit activities in Bitcoin transaction networks often manifest as coordinated multi-hop transaction chains rather than isolated interactions. These activities differ in legitimate transactions and exhibit properties like reactivation of dormant addresses and closures of newly created addresses\cite{heidarinia2014intelligent}. In particular, laundering, ransomware, and mixing behaviors frequently involve splitting funds across multiple intermediary addresses, producing fan-out transaction patterns \cite{tovanich2022pattern, zola2025topological}.

\begin{table*}[t]
\centering
\caption{Comparison of prior work by dataset, models, metrics, and embedding geometry.}
\label{tab:prior-work-check}
\renewcommand{\arraystretch}{1.2}
\begin{tabular}{lccccccccccc}
\toprule
 & \multicolumn{2}{c}{\textbf{Dataset}}
 & \multicolumn{4}{c}{\textbf{Models}}
 & \multicolumn{3}{c}{\textbf{Metrics}}
 & \multicolumn{2}{c}{\textbf{Geometry}} \\
\cmidrule(lr){2-3}
\cmidrule(lr){4-7}
\cmidrule(lr){8-10}
\cmidrule(lr){11-12}
\textbf{Paper}
& New BTC Graph
& Elliptic
& GCN
& GraphSAGE
& GAT
& HyperGNN
& Precision
& Recall
& F1
& Euclidean
& Hyperbolic \\
\midrule

\cite{weber2019anti}
& \xmark & \cmark
& \cmark & \xmark & \xmark & \xmark
& \cmark & \cmark & \cmark
& \cmark & \xmark \\

\cite{lo2023inspection}
& \xmark & \cmark
& \cmark & \cmark & \xmark & \xmark
& \cmark & \cmark & \cmark
& \cmark & \xmark \\

\cite{bellei2024shape}
& \xmark & \cmark
& \cmark & \xmark & \xmark & \xmark
& \cmark & \cmark & \cmark
& \cmark & \xmark \\

\cite{schnoering2025bitcoin}
& \cmark & \xmark
& \cmark & \xmark & \xmark & \xmark
& \cmark & \cmark & \cmark
& \cmark & \xmark \\

\cite{CHAMIGCN}
& \xmark & \xmark
& \cmark & \xmark & \xmark & \cmark
& \xmark & \xmark & \cmark
& \xmark & \cmark \\

\textbf{This work}
& \cmark & \xmark
& \cmark & \cmark & \cmark & \cmark
& \cmark & \cmark & \cmark
& \cmark & \cmark \\

\bottomrule
\end{tabular}
\end{table*}

\subsection{Graph Neural Networks in Transaction and Social Systems}

Prior to the widespread adoption of GNNs, early studies had demonstrated that blockchain transaction graphs exhibit distinct structural patterns between different entities and activities. Weber \emph{et al.} pioneered the use of graph-based analysis for cryptocurrency transactions in \cite{weber2019anti} where he showed that handcrafted topological features can capture transactional patterns differences between legitimate and fraud  transactions. Similarly, Akcora \emph{et al.} leveraged local topological features extracted from the Bitcoin transaction graph to detect ransomware-related addresses, demonstrating that structured payment patterns alone can effectively distinguish malicious behavior without relying on deep neural architectures \cite{AKCORA}.

Building upon these insights, subsequent work began incorporating graph representation learning to automatically capture behavioral patterns from transaction graphs. Zhengjie \emph{et al.} proposed BAClassifier, a GNN-based framework that models each Bitcoin address as a chronological transaction graph and learns address-level embeddings for behavior classification\cite{BAClassifier}. Their approach employs graph node compression and structure augmentation to construct unified address graphs and avoids reliance on off-chain information by directly leveraging on-chain transaction topology. More recently, GNN architectures have been extended to handle heterogeneity and temporal dynamics in blockchain systems. Lin \emph{et al.} introduced a Heterogeneous Graph Transformer Network (HGTN) for smart contract anomaly detection on the Ethereum platform \cite{LIU}. Zhang \emph{et al.} proposed DA-HGNN, a hybrid phishing detection framework that integrates data augmentation, temporal sequence modeling, and graph representation learning to identify phishing accounts in Ethereum transaction networks \cite{CHEN}.

\subsection{Hyperbolic Representation Learning}

Hyperbolic geometry has emerged as an effective representation space for data exhibiting latent hierarchical or tree-like structure. Unlike Euclidean space, which has polynomial volume growth, hyperbolic space expands exponentially with radius, allowing it to embed branching and hierarchical relationships with low distortion. Nickel and Kiela first demonstrated the advantages of hyperbolic embeddings using the Poincar\'e ball model for learning symbolic hierarchies \cite{nickel2017poincare}. Subsequently, Sala et al. formalized the tradeoffs between precision and dimensionality by introducing a reconstruction that achieves low distortion for trees without optimization\cite{sala2018representation}.

Building on these, hyperbolic representation learning has been extended to graph-structured data. Early approaches focused on embedding nodes or entire graphs into hyperbolic space for tasks like link prediction and node classification showing improved performance on graphs with hierarchical organization \cite{liu2019hyperbolic,CHAMIGCN}. More recent work has integrated hyperbolic geometry into graph neural networks, enabling tasks and operations like message passing and neighborhood aggregation under non-Euclidean metrics. Fully hyperbolic GNNs define aggregation, attention, and nonlinear transformations using Riemannian operations which preserves geometric consistency but introduces substantial computational and optimization complexity \cite{CHAMIGCN}.

To address these challenges, hyperbolic GNNs tangent space formulations have been proposed, in which node representations are mapped to a Euclidean tangent space via logarithmic maps, processed using standard neural operations, and projected back into hyperbolic space through exponential maps \cite{yang2022hyperbolic}. These approaches significantly improve numerical stability and scalability while retaining the representational benefits of hyperbolic geometry, making them suitable for large-scale graph learning.

These evaluations of hyperbolic GNNs have mostly focused on benchmark citation graphs, social networks, knowledge graphs\cite{dai2021hyperbolic, lee2023node}. Despite these advances, hyperbolic graph neural networks have not been systematically studied in blockchain transaction networks or any large scale financial transaction systems. Existing studies typically adopt fixed neighborhood structures or full-graph training and do not explicitly examine how changing the sampled neighborhood depth interacts with embedding geometry in transaction graphs. 

Table~\ref{tab:prior-work-check} provides a structured summary of representative studies, comparing datasets, model families, evaluation metrics, and embedding geometries. It contextualizes our experimental design and highlights how our work fits within the existing literature.

\begin{figure*}[t]
    \centering
    \includegraphics[width=1.05\textwidth, trim={0.5cm 17cm 0 0}, clip]{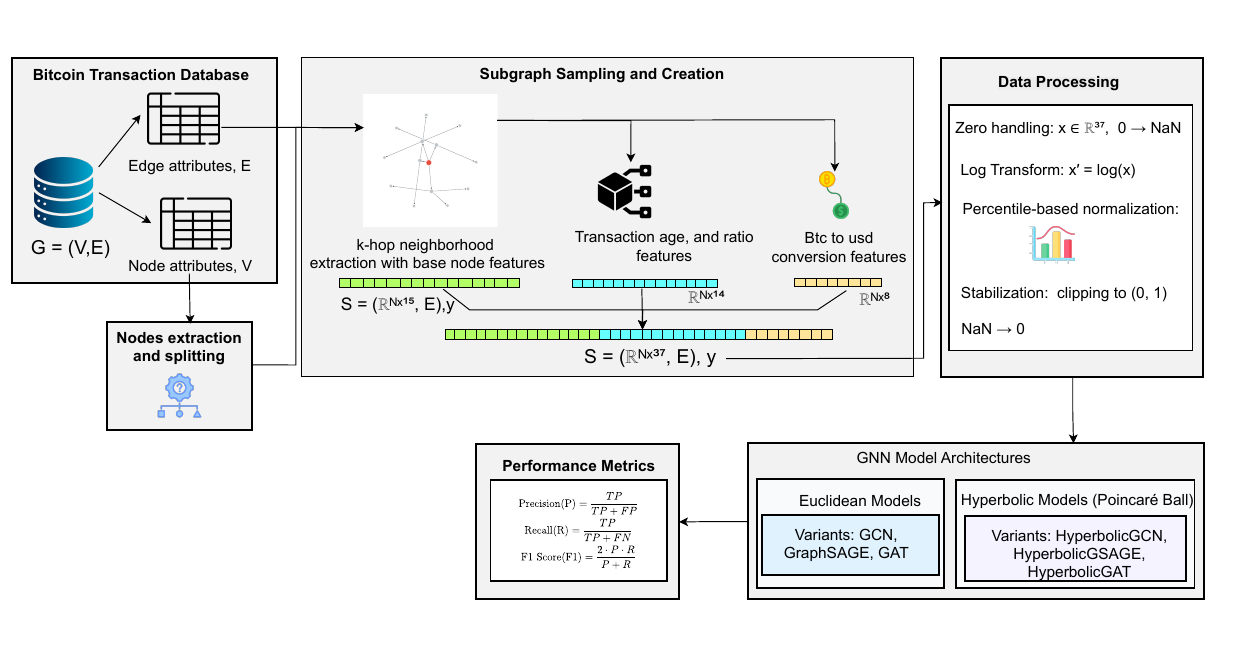} 
    \captionsetup{justification=centering} 
    \caption{Workflow for Bitcoin Address Classification.}
    \label{fig:address_classification_workflow}
\end{figure*}

\section{Methodology}

Figure~\ref{fig:address_classification_workflow} presents the complete workflow for Bitcoin address classification used in this study.

\subsection{Bitcoin Transaction Database}
We use the large-scale Bitcoin transaction graph dataset from Schnoering and Vazirgiannis~\cite{schnoering2025bitcoin}, comprising approximately 252 million nodes and 785 million edges. The graph G = (V,E) has nodes representing Bitcoin addresses and directed edges representing transactions. We focus on 34,000 labeled nodes across seven entity classes: Exchange, Mining, Gambling, Ponzi, Individual, Ransomware, and Bet. Compared to previously used datasets such as Elliptic, which contains approximately 200,000 nodes but only two class labels (licit and illicit), this dataset provides a richer label taxonomy across a substantially larger graph allowing for a granular evaluation on the effect of embedding geometry. The scale and diversity of this dataset makes it one of the most informative available choice for studying the interaction between neighborhood depth and embedding geometry in Bitcoin transaction networks.

We split the labeled nodes into train (40\%), validation (30\%), and test (30\%) sets using stratified sampling.  Due to severe class imbalance, random oversampling is applied only to the training split such that each class contains 300 samples while the validation and test splits remain unchanged.

\subsection{Subgraph Sampling and Creation}

Due to the large-scale nature of the graph, we construct ego-centric subgraphs around labeled nodes. For each labeled seed node $v$, we construct a sampled subgraph $G_v^{(k)} = (V_v^{(k)}, E_v^{(k)})$, where $V_v^{(k)}$ contains nodes within $k$ hops of $v$ and $E_v^{(k)}$ includes all edges among sampled nodes.

\subsubsection{Sampling Strategy}
We evaluate two neighborhood depths to analyze how increasing transactional context affects representation learning in different embedding geometries:
\begin{itemize}
    \item \textbf{Depth-2:} Two hops with fan-out \{1:5, 2:10\}
    \item \textbf{Depth-3:} Three hops with fan-out \{1:5, 2:10, 3:8\}
\end{itemize}

Unlike~\cite{schnoering2025bitcoin}, which employ neighborhood buffering and multiple samples per node for data augmentation, we construct a single fixed subgraph per labeled node. This controlled approach ensures that performance differences can be attributed to geometric embedding choices rather than sampling variance, allowing us to isolate the impact of hyperbolic versus Euclidean representations. Since neighborhood expansion is based on fixed fan-out sampling rather than full k-hop extraction, increasing subgraph depth reflects increased sampled transactional exposure rather than complete structural expansion.

\subsubsection{Feature Construction}
We based our feature construction pipeline on the approach proposed by Schnoering and Vazirgiannis~\cite{schnoering2025bitcoin}, adapting it to our computational setup and analysis objectives. Our focus was to prepare node features to capture transactional behavior while isolating the impact of different graph embedding geometries rather than introducing additional feature engineering choices.

We use the original transaction features of the nodes along with additional derived features including average amounts sent and received, in-degree and out-degree ratios, cluster composition ratios, node age, and activity rates normalized by node age. Following the reconstruction of the original dataset, all value-type features are converted from satoshi to USD. Historical BTC/USD conversion rates are used to compute the median satoshi-to-USD price over the active lifetime of each node to ensure meaningful comparison of transaction values across time periods.

\subsection{Data Processing}

The features exhibited a probability density that followed a power law distribution, except at very large values. Following~\cite{schnoering2025bitcoin}, we apply a five-stage normalization pipeline to enhance learning.

\paragraph{Zero Handling for Logarithmic Stability.} 
To facilitate the logarithmic transformation of features following a power-law distribution, zero values are temporarily replaced with NaNs. This prevents numerical divergence ($\log(0)$) and ensures the transformation is applied only to the positive transactional range. These values are later restored to the distribution's origin during the final processing stage, mapping the absence of activity to the baseline of the normalized feature space.

\paragraph{Log Transform: $x' = \log(x)$.}
Logarithmic transformation is applied to address the power law distribution of features, which can span several orders of magnitude in transaction amounts.

\paragraph{Percentile-Based Normalization.}
Each feature is normalized using its minimum value $q_{0\%}$ and the 95th percentile $q_{95\%}$ through the transformation:
\begin{equation}
x \rightarrow \frac{\log(x) - \log(q_{0\%})}{\log(q_{95\%}) - \log(q_{0\%})}
\end{equation}
For value-type features, smaller values did not adhere well to a power law distribution. Therefore, we substituted the minimum value with the 5th percentile $q_{5\%}$ in the normalization formula.

\paragraph{Stabilization: Clipping to $(0, 1)$.}
Normalized features are clipped to the range $[0, 1]$ to prevent extreme values from dominating the feature space.

\paragraph{Missing Value Replacement: $\text{NaN} \rightarrow 0$.}
Remaining NaN values (from the initial zero handling) are replaced with 0 to produce complete feature vectors.

All normalization constants are computed exclusively on the training split to prevent data leakage. These same constants are then applied to validation and test splits, ensuring that model evaluation reflects true generalization performance on unseen data.

\begin{figure*}[htbp]
    \centering
    \includegraphics[width=1\textwidth, trim={0 18cm 0 1cm}, clip]
    {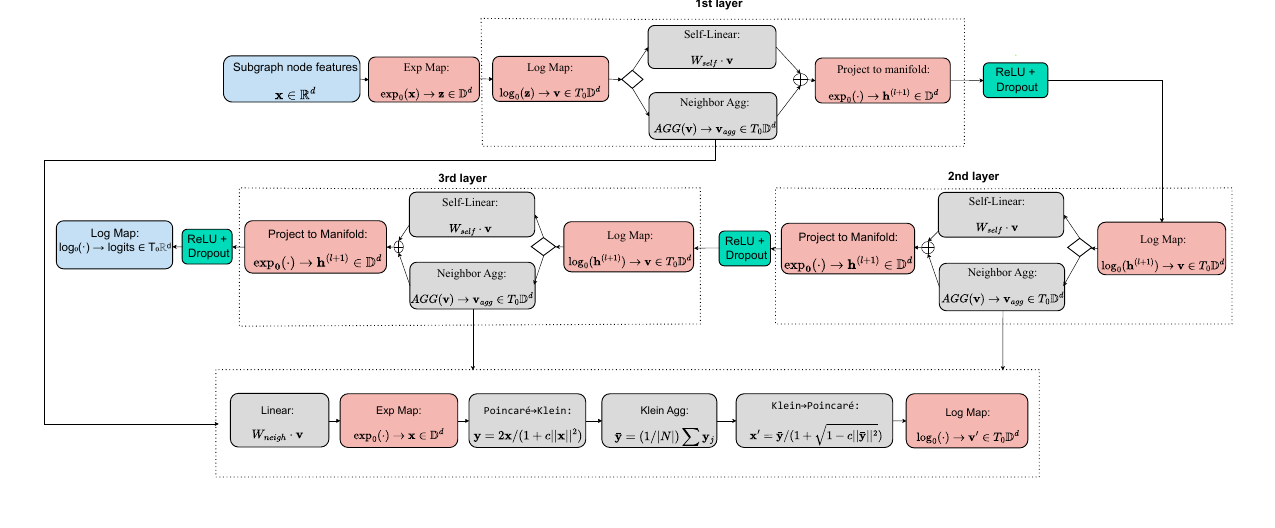}
    \caption{HyperbolicSAGE architecture showing the computational flow across layers.}
    \label{fig:hyperbolic_sage_architecture}
\end{figure*}

\subsection{GNN Model Architectures}

We adopt three canonical GNN architectures-GCN, GraphSAGE, and GAT-in two geometric spaces: Euclidean space and Hyperbolic space. These architectures represent different aggregation strategies and serve as strong baselines to assess the impact of different geometry. All architectures are evaluated using both 2-layer and 3-layer configurations on subgraphs of depth=2 and 3. The key distinction between Euclidean and hyperbolic variants lies in the embedding geometry and how operations are distributed across geometric spaces.

\subsubsection{GNNs in Euclidean Space} 
All models operate in Euclidean space $\mathbb{R}^d$ and update node representations through neighborhood message passing. Let $\mathbf{H}^{(k)} \in \mathbb{R}^{n \times d}$ denote the node feature matrix at layer $k$, where $n$ is the number of nodes in the subgraph and $d$ is the feature dimension. In all experiments, we fix the embedding dimension $d=256$ for both Euclidean and hyperbolic models. $\mathcal{E}$ denotes the edge set of the subgraph.

\paragraph{\textbf{GCN}} Our implementation uses PyTorch Geometric's GCNConv layer, which performs symmetric normalization based on node degrees. Following each GCNConv layer except the final one, we apply ReLU activation and dropout (rate 0.1):
\begin{equation}
\mathbf{H}^{(k)} = \text{Dropout}(\text{ReLU}(\text{GCNConv}(\mathbf{H}^{(k-1)}, \mathcal{E})), 0.1)
\end{equation}
The final GCNConv layer outputs logits directly without activation, from which we extract predictions only for seed nodes.

\paragraph{\textbf{GraphSAGE}} We use SAGEConv with the default mean aggregator, which concatenates each node's features with its aggregated neighborhood. Both 2-layer and 3-layer configurations follow the same pattern:
\begin{equation}
\mathbf{H}^{(k)} = \text{Dropout}(\text{ReLU}(\text{SAGEConv}(\mathbf{H}^{(k-1)}, \mathcal{E})), 0.1)
\end{equation}
This concatenation operation preserves node identity while incorporating neighborhood context, which is important for distinguishing seed addresses from their transaction neighbors.

\paragraph{\textbf{GAT}} Our GAT implementation uses a custom dual-transformation attention layer rather than the standard single-transformation variant. Each layer processes target and neighbor nodes through separate weight matrices ($\mathbf{W}_1$ and $\mathbf{W}_2$) before computing attention coefficients. We employ 8 attention heads in all intermediate layers (32 dimensions per head) and a single head in the final layer. For a 3-layer configuration:
\begin{equation}
\begin{aligned}
\mathbf{H}^{(k)} =
\text{Dropout}\big(&
\text{ReLU}(
\text{GAT}(\mathbf{H}^{(k-1)}, \mathcal{E}, \text{heads}=8)
), \\
&0.1
\big), \quad k = 1,2
\end{aligned}
\end{equation}

All models use Adam optimizer with learning rate 0.001 and are trained to classify only the seed nodes in each subgraph, ignoring the surrounding transaction context during loss computation.

\subsubsection{GNNs in Hyperbolic Space (Poincaré Ball)}
Bitcoin transaction networks exhibit locally tree-like expansion patterns. Hyperbolic space, with constant negative curvature and exponential volume growth, naturally accommodates these rapidly expanding structures with lower distortion than Euclidean space. We use the Poincaré ball model:
\begin{equation}
\mathbb{B}_c^d = \left\{ \mathbf{x} \in \mathbb{R}^d \mid c \lVert \mathbf{x} \rVert^2 < 1 \right\}
\end{equation}
where $c > 0$ controls curvature.
For all hyperbolic architectures, the final layer omits ReLU and produces logits in tangent space.

All hyperbolic models coordinate operations across multiple geometric spaces using the exponential map to project tangent vectors to the Poincaré ball:
\begin{equation}
\exp_0^c(\mathbf{v}) = \tanh(\sqrt{c}\|\mathbf{v}\|) \frac{\mathbf{v}}{\sqrt{c}\|\mathbf{v}\|}
\end{equation}
and the logarithmic map to return points to tangent space for linear transformations:
\begin{equation}
\log_0^c(\mathbf{x}) = \tanh^{-1}(\sqrt{c}\|\mathbf{x}\|) \frac{\mathbf{x}}{\sqrt{c}\|\mathbf{x}\|}
\end{equation}

For neighborhood aggregation, we perform mean aggregation in the Klein model:
\begin{equation}
\text{KleinAgg}(\mathbf{H}, \mathcal{N}(v))
= \frac{1}{|\mathcal{N}(v)|} \sum_{u \in \mathcal{N}(v)} \mathbf{h}_u
\end{equation}
where aggregation is performed in Klein coordinates rather than Poincaré 
coordinates.

\paragraph{\textbf{HyperbolicGCN}} The input features are first projected to the Poincaré ball:
\begin{equation}
\mathbf{H}^{(0)} = \exp_0^c(\mathbf{X})
\end{equation}
where $\mathbf{X}$ are the Euclidean input features. For each layer, features are aggregated in Klein space then processed through tangent space transformations:
\begin{equation}
\mathbf{h}_v^{\text{agg}} = \text{KleinAgg}(\mathbf{H}^{(k-1)}, \mathcal{N}(v))
\end{equation}
\begin{equation}
\mathbf{H}^{(k)} = \exp_0^c(\text{ReLU}(\mathbf{W}^{(k)} \log_0^c(\mathbf{H}_{\text{agg}})))
\end{equation}
where $\text{KleinAgg}$ performs mean aggregation in the Klein model. The final layer omits ReLU and produces logits in tangent space.

\paragraph{\textbf{HyperbolicSAGE}} We adapt GraphSAGE's dual transformation to hyperbolic space. After projecting inputs to Poincaré space, each layer applies separate transformations to self and neighbor features:
\begin{equation}
\mathbf{H}_1 = \mathbf{W}_1 \log_0^c(\mathbf{H}^{(k-1)})
\end{equation}
\begin{equation}
\mathbf{H}_2 = \text{KleinAgg}(\mathbf{W}_2 \log_0^c(\mathbf{H}^{(k-1)}), \mathcal{N}(v))
\end{equation}
The features are then combined and mapped back:
\begin{equation}
\mathbf{H}^{(k)} = \exp_0^c(\text{ReLU}(\mathbf{H}_1 + \mathbf{H}_2))
\end{equation}
Both 2-layer and 3-layer variants include self-loops by default. Figure~\ref{fig:hyperbolic_sage_architecture} illustrates the computational flow for HyperbolicSAGE.

\paragraph{\textbf{HyperbolicGAT}}
We implement a dual-transformation distance-based attention mechanism in hyperbolic space, where attention coefficients are computed from the negative hyperbolic distance between separately transformed target and neighbor embeddings

After projecting to Poincaré space, each attention head applies dual transformations to node features:
\begin{equation}
\begin{aligned}
\mathbf{h}_{v}' &= \exp_0^c(\mathbf{W}_1\log_0^c(\mathbf{h}_v^{(k-1)})), \\
\mathbf{h}_{u}' &= \exp_0^c(\mathbf{W}_2\log_0^c(\mathbf{h}_u^{(k-1)}))
\end{aligned}
\end{equation}

where $\mathbf{W}_1$ and $\mathbf{W}_2$ are separate weight matrices for target and neighbor nodes respectively. Attention coefficients are computed as negative hyperbolic distances between these transformed features:
\begin{equation}
\alpha_{vu} = \frac{\exp(-d_{\mathbb{H}}(\mathbf{h}_{v}', \mathbf{h}_{u}'))}{\sum_{k \in \mathcal{N}(v)} \exp(-d_{\mathbb{H}}(\mathbf{h}_{v}', \mathbf{h}_{k}'))}
\end{equation}
Features are aggregated in tangent space using attention weights and combined with self features:
\begin{equation}
\mathbf{H}_{\text{agg}} = \sum_{u \in \mathcal{N}(v)} \alpha_{vu} \mathbf{W}_2 \log_0^c(\mathbf{h}_u^{(k-1)})
\end{equation}
\begin{equation}
\mathbf{H}^{(k)} = \exp_0^c(\text{ReLU}(\mathbf{W}_1 \log_0^c(\mathbf{H}^{(k-1)}) + \mathbf{H}_{\text{agg}}))
\end{equation}
We use 8 attention heads in intermediate layers (32 dimensions per head) and a single head for the final layer.

All hyperbolic models use Adam optimizer with learning rate 0.001 and classify only seed nodes. Curvature $c$ is selected from $\{0.1, 0.3, 0.5, 0.75, 1.0, 1.25, 1.5\}$ based on validation performance.

\subsection{Performance Metrics}

To evaluate the performance of all models, we considered
different performance metrics, including precision, recall, and F1-score and evaluated on both per-class and aggregate levels. These metrics provide a comprehensive view of the model's performance over varied neighborhoods. Due to the huge class imbalance, macro-averaged F1-score was the primary evaluation metric to ensure that the performance gains are not dominated by the classes with larger samples. The mathematical definitions of these performance metrics are presented in Figure 2 which illustrates how the evaluation was conducted.

\begin{table*}[t]
\centering
\caption{Node classification performance on depth-two subgraphs (macro-averaged). Columns correspond to different GNN architectures; rows indicate model depth and evaluation metric. The highest value in each row is highlighted in bold.}
\label{tab:depth2-wide}
\begin{tabular}{llcccccc}
\toprule
Layers & Metric & GCN & HGCN & GraphSAGE & HGraphSAGE & GAT & HGAT \\
\midrule
\multirow{3}{*}{2-layer}
 & Precision & 0.68 & 0.67 & 0.81 & \textbf{0.82} & 0.79 & 0.82 \\
 & Recall    & 0.49 & 0.55 & 0.67 & \textbf{0.71} & 0.65 & \textbf{0.71} \\
 & Macro-F1  & 0.55 & 0.59 & 0.72 & \textbf{0.76} & 0.71 & 0.75 \\
\midrule
\multirow{3}{*}{3-layer}
 & Precision & 0.67 & 0.75 & 0.82 & \textbf{0.87} & 0.79 & 0.86 \\
 & Recall    & 0.54 & 0.57 & 0.67 & \textbf{0.75} & 0.69 & 0.72 \\
 & Macro-F1  & 0.59 & 0.64 & 0.73 & \textbf{0.80} & 0.73 & 0.78 \\
\bottomrule
\end{tabular}
\end{table*}

\begin{table*}[t]
\centering
\caption{Node classification performance on depth-three subgraphs (macro-averaged). Columns correspond to different GNN architectures; rows indicate model depth and evaluation metric. The highest value in each row is highlighted in bold.}
\label{tab:depth3-wide}
\begin{tabular}{llcccccc}
\toprule
Layers & Metric & GCN & HGCN & GraphSAGE & HGraphSAGE & GAT & HGAT \\
\midrule
\multirow{3}{*}{2-layer}
 & Precision & 0.68 & 0.74 & 0.84 & 0.83 & 0.77 & \textbf{0.85} \\
 & Recall    & 0.52 & 0.58 & \textbf{0.72} & 0.71 & 0.70 & 0.71 \\
 & Macro-F1  & 0.58 & 0.64 & \textbf{0.77} & 0.76 & 0.73 & \textbf{0.77} \\
\midrule
\multirow{3}{*}{3-layer}
 & Precision & 0.71 & 0.74 & 0.84 & \textbf{0.88} & 0.85 & 0.86 \\
 & Recall    & 0.60 & 0.63 & 0.73 & 0.76 & 0.73 & \textbf{0.77} \\
 & Macro-F1  & 0.64 & 0.68 & 0.78 & \textbf{0.81} & 0.78 & \textbf{0.81} \\
\bottomrule
\end{tabular}
\end{table*}

\begin{table*}[t]
\centering
\caption{Per-class node classification performance (macro-averaged metrics) for selected Euclidean and hyperbolic GNNs. Bold indicates the best value per class; ties are all bolded.}
\label{tab:per-class}
\begin{tabular}{lccccccc}
\toprule
Class & Metric & GCN & HGCN & GraphSAGE & HGraphSAGE & GAT & HGAT \\
\midrule
BET & Precision & 0.96 & 0.96 & 0.97 & \textbf{0.98} & 0.97 & \textbf{0.98} \\
    & Recall    & 0.95 & 0.96 & \textbf{0.99} & \textbf{0.99} & \textbf{0.99} & \textbf{0.99} \\
    & F1        & 0.95 & 0.96 & 0.98 & \textbf{0.99} & 0.98 & \textbf{0.99} \\
\midrule
EXCHANGE & Precision & 0.73 & 0.67 & 0.83 & \textbf{0.85} & 0.84 & 0.82 \\
         & Recall    & 0.40 & 0.53 & 0.62 & 0.67 & 0.63 & \textbf{0.70} \\
         & F1        & 0.51 & 0.59 & 0.71 & 0.75 & 0.72 & \textbf{0.76} \\
\midrule
GAMBLING & Precision & 0.81 & 0.77 & 0.83 & \textbf{0.88} & 0.86 & 0.86 \\
         & Recall    & 0.59 & 0.67 & 0.75 & 0.77 & 0.74 & \textbf{0.78} \\
         & F1        & 0.68 & 0.72 & 0.79 & \textbf{0.82} & 0.80 & \textbf{0.82} \\
\midrule
INDIVIDUAL & Precision & 0.92 & 0.93 & 0.96 & 0.96 & 0.96 & \textbf{0.96} \\
           & Recall    & 0.97 & 0.96 & 0.98 & \textbf{0.99} & 0.98 & 0.98 \\
           & F1        & 0.94 & 0.95 & 0.97 & \textbf{0.97} & 0.97 & \textbf{0.97} \\
\midrule
MINING & Precision & 0.59 & 0.63 & 0.79 & \textbf{0.88} & 0.86 & 0.80 \\
       & Recall    & 0.37 & 0.42 & 0.63 & 0.66 & 0.59 & \textbf{0.68} \\
       & F1        & 0.45 & 0.50 & 0.70 & \textbf{0.75} & 0.70 & 0.73 \\
\midrule
PONZI & Precision & 0.47 & 0.62 & 0.74 & \textbf{0.84} & 0.70 & 0.81 \\
      & Recall    & 0.53 & 0.51 & 0.58 & \textbf{0.63} & 0.56 & 0.61 \\
      & F1        & 0.50 & 0.56 & 0.65 & \textbf{0.72} & 0.62 & 0.70 \\
\midrule
RANSOMWARE & Precision & 0.53 & 0.59 & 0.77 & 0.77 & 0.73 & \textbf{0.81} \\
           & Recall    & 0.41 & 0.37 & 0.59 & \textbf{0.63} & \textbf{0.63} & 0.61 \\
           & F1        & 0.46 & 0.46 & 0.67 & 0.69 & 0.68 & \textbf{0.70} \\
\bottomrule
\end{tabular}
\end{table*}

% =============================
% 5. Experimental Results
% =============================
\section{Experimental Results}

This section presents a complete comparison of Euclidean and hyperbolic graph neural networks on the Bitcoin transaction graph. By keeping the subgraph construction and preprocessing identical along with the comparable architecture of all models, we tried to isolate the effects of underlying geometry and how the geometry affects the performance as the neighborhood sampling radius increases.

\subsection{Effect of Network Depth at Fixed Subgraph Depth}

We analyze the interaction between architectural depth and neighborhood expansion by comparing two- and three-layer GNNs operating on subgraphs of fixed depth. This allows us to isolate the effect of network depth from that of subgraph sampling depth.

\subsubsection{Subgraph Depth Two}
Table~\ref{tab:depth2-wide} reports node classification performance on two-hop neighborhood subgraphs for both two- and three-layer GNNs. For these subgraphs, upon using shallower depth models we found that the best Euclidean models perform comparably to their hyperbolic counterparts. Upon adding a third layer, the hyperbolic models saw significant gains with HGraphSAGE achieving a macro-F1 score of 0.80 compared to 0.73 for the best Euclidean models.

\subsubsection{Subgraph Depth Three}
We next increased the receptive field for all seed nodes and carried the same task on the subgraphs extracted at depth three. Table~\ref{tab:depth3-wide} summarizes the corresponding results on subgraphs of depth three. Similar to the subgraphs at two depth, hyperbolic models show significant advantages for 3-layer architectures, with HGraphSAGE achieving the best performance of 0.81 macro-F1.

Comparing Tables~\ref{tab:depth2-wide} and~\ref{tab:depth3-wide}, we observe that hyperbolic architectures maintain a consistent performance advantage in macro-F1 across different depth configurations. The primary benefit appears to come from the combination of hyperbolic geometry with deeper models, which consistently outperforms all other models. Notably, with sufficiently deeper models, the advantage of hyperbolic geometry is already present at moderate depths of subgraphs and does not require extremely large neighborhoods to become apparent.

\subsection{Per-Class Performance Analysis}

To better understand how geometry affects model behavior, we also examine performance at the level of individual classes. 
Hyperbolic architectures show consistent improvements across most classes rather than having gains concentrated for specific entity classes. HGraphSAGE in particular achieves the highest recall for nearly all classes, highlighting that the benefit of hyperbolic geometry is not limited to specific classes but expands to all entity transaction subgraph representations.

For high-frequency classes such as Bet and Individual, all models already achieve a strong baseline performance (F1 $>$ 0.91). Because these categories make up the largest chunk of the subgraph samples, their structural and feature patterns are more easily distinguishable and the performance differences between Euclidean and hyperbolic models are minimal. The high performance by the baseline models leaves limited room for improvement.

However, for more challenging classes such as Exchange, Gambling, Mining, Ponzi, and Ransomware, hyperbolic models show more meaningful gains over their best Euclidean counterparts. The most notable improvement is observed for Ponzi, where HGraphSAGE improves precision from 0.74 to 0.84 and F1 from 0.65 to 0.72 over the best Euclidean baseline. For Gambling, HGAT achieves the best recall of 0.78 compared to 0.75 for GraphSAGE. For Mining and Exchange, hyperbolic models show modest but consistent F1 improvements of 0.05 and 0.04 respectively over the best Euclidean baselines. These improvements suggest that hyperbolic models are particularly helpful for entity types whose transactional patterns are more structurally ambiguous and harder to separate in Euclidean space.

From an application perspective, recall on rare categories such as Ponzi and Ransomware is particularly important in fraud detection. Although the gains on these classes are meaningful, they are comparable in magnitude to improvements observed for other structurally challenging categories. This suggests that the primary strength of hyperbolic models lies in their ability to improve performance broadly across moderately difficult classes, rather than providing isolated benefits for explicitly fraudulent entities.

\subsection{Optimization Behavior Under Curvature and Learning Rate}

We performed a grid search over joint combinations of learning rates and curvature to identify the most effective hyperparameters for tangent-space hyperbolic GNNs. Beyond hyperparameter tuning, this analysis reveals fundamental insights about how curvature shapes the optimization landscape and interacts with learning rate selection.

Our analysis revealed that curvature has a dominant influence on the effective optimization with very low curvature yielding under trained models regardless of the learning rate. At lower curvature values ($c < $ 0.5), the models consistently converge to poor local minima, achieving inadequate performance regardless of learning rate tuning. However, sufficient curvature ($c \ge 1.0$) transforms the optimization landscape, enabling robust performance across huge range of learning rates. 

We identified an transition region around c = 0.75–1.25, below which optimization remains fragile. Learning rate plays a secondary role but with observable effects: insufficient curvatures and low learning rates result in under trained models while at sufficient curvature even extremely low learning rate achieve strong performance.  For instance, at c = 1.25, validation F1 varied by only 0.07 points across all learning rates tested (0.748-0.821), whereas at c = 0.10, the same learning rate range produced a 0.59 point variation (0.117-0.708). At low curvatures, while higher learning rates can partially compensate for geometric inadequacy, it is unreliable with subpar performance.

\begin{figure}[h]
    \centering
    \includegraphics[width=0.9\linewidth]{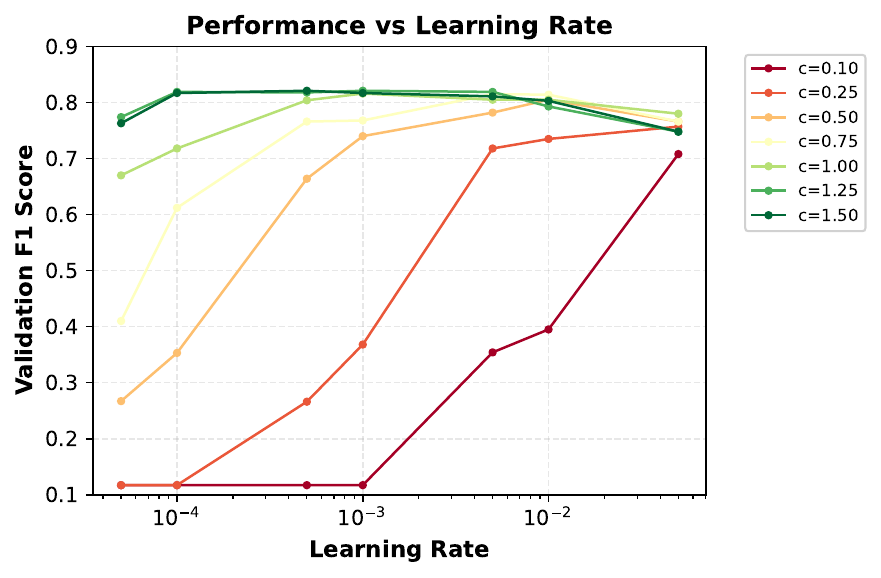}

    \captionsetup{justification=justified, margin=0pt, width=\linewidth, singlelinecheck=false}
    \caption{Macro-F1 versus learning rate for different curvature ($c$) in tangent-space Hyperbolic GraphSAGE. Low curvature values at low to moderate learning rates produce large regions of convergent but uninformative solutions. Moderate to high curvature values achieve effective performance across a wider learning-rate range, highlighting the importance of joint learning-rate and curvature selection.}
    \label{fig:graphsage-curvature-lr}
\end{figure}

Figure~\ref{fig:graphsage-curvature-lr} illustrates these patterns for Hyperbolic GraphSAGE. Each curve shows validation macro-F1 across learning rates for a fixed curvature value. Low curvature values (red/orange lines) show steep performance increase with respect to learning rate, particularly at low to moderate values. The gap between c = 0.50 and c = 0.75 at LR = 0.0001 is substantial (F1 = 0.353 vs. 0.612). High curvature values (green lines, $x \ge 5$) show relatively flat performance across the learning rate spectrum, indicating reduced sensitivity to this hyperparameter choice.

\subsection{Theoretical Interpretation of Embedding Geometry}

The advantage of hyperbolic models across both sampling depths can be understood in terms of the interaction between embedding geometry and the structure induced by our sampling strategy. The subgraphs were constructed with a fixed fan-out at each hop, inducing a tree-like expansion pattern with each depth.

In Euclidean space, volume grows polynomially with radius. Hence, representing neighborhoods with such branching patterns places pressure on the embedding space, leading to crowding effects where distinct multi-hop transaction paths become difficult to separate. Hyperbolic space, by contrast, provides a geometry that may be better aligned with these expansion patterns, offering greater capacity to preserve separation between branches.

The marginal performance differences observed at different depths may also reflect the limitations of fixed fan-out sampling. Since sampled subgraphs represent only a subset of the transaction neighborhood, increasing the hop count does not necessarily introduce richer structural information but can even misrepresent it. As a result, the relative advantage between hyperbolic and Euclidean models remains stable or not substantial across depths.

\subsection{Complexity and Computational Considerations}
From a complexity perspective, the GCN architecture scales as $O(|\mathcal{E}|CHF)$, where $|\mathcal{E}|$ represents the cardinality of the edge set. In our implementation, node features are treated as 1D vectors ($H=1$) with a fixed hidden dimensionality $d$, simplifying the complexity for input ($C=d$) and output ($F=d$) layers to $O(E \cdot d^2)$. The additional operations required for hyperbolic tangent-space mappings are \textit{element-wise} and scale as $O(V \cdot d)$. Since $V \cdot d \ll E \cdot d^2$ in our sampled transaction subgraphs, the hyperbolic transformation maintains the same asymptotic complexity as the Euclidean baseline.

All models were trained and evaluated on a 4-node cluster, where each node was equipped with an NVIDIA GeForce RTX 5070 Ti GPU with 16 GB of memory. We observed that hyperbolic models incurred slightly higher training time and memory usage compared to Euclidean baselines, primarily due to the transcendental function evaluations in the logarithmic and exponential mappings. However, the tangent-space formulation rendered these models practical for large-scale financial transaction graphs. We found that careful selection of the learning rate and curvature was critical for effective optimization under these resource constraints. Subgraph construction was performed as a one-time, offline preprocessing step on a CPU-based SLURM cluster using array jobs to parallelize train, validation and test splits. Each split was processed independently on a single node with 10 CPU workers and 64 GB RAM, extracting fixed ego-centric subgraphs using multi-hop neighborhood sampling at the specified depth. Due to the scale of the full Bitcoin transaction graph, subgraph generation relied on disk-backed, multi-hop traversal of a PostgreSQL-backed graph store.

\section{Conclusion}
We conducted an extensive review of hyperbolic and Euclidean graph neural networks to classify entities in the Bitcoin network and to model Bitcoin transaction networks. We found that hyperbolic representations were significantly better than Euclidean representations at capturing branching patterns, especially when using deep architectures. These improvements were most pronounced in cases where Euclidean aggregation was poor at capturing the transactional patterns of the entity type being classified. Hyperbolic representation optimization success was largely determined by curvature. In particular, we found that values of $c \geq 1.0$ enabled stable optimization results across a range of learning rates. While Euclidean GNNs provide adequate results for many blockchain applications, we must consider geometric properties of the embedding space when developing methods for detecting entities or fraudulent behavior within cryptocurrency networks.

\section{LIMITATIONS AND FUTURE WORK}

The objective of our study was to investigate how the geometric space of embeddings affects the representation of the neighborhood in a way that can be compared to the effects of the number of hops (i.e., neighborhood depth) in a controlled setting of a large-scale Bitcoin transaction network. Therefore, we limited ourselves to only examining static transaction graphs, and we used fixed-depth and fixed-fanout-parameters for ego-centric subgraph sampling to enable reproducibility of the comparisons. We didn't model temporal dynamics or dependencies greater than those contained within the sampled neighborhoods. Further, our hyperbolic models used a tangent-space formulation with a fixed curvature value to prioritize numerical stability and scalability. As such, future studies could extend this research to include temporal modeling, adaptive or mixed-geometry representations, and/or different neighborhood sampling methods, as well as other blockchain or financial transaction systems, and/or to include other downstream forensic tasks (e.g., early fraud detection and/or risk assessment).

% =============================
% References
% =============================
\bibliographystyle{IEEEtran}
\bibliography{references}

\end{document}